\pdfoutput=1
\documentclass[11pt]{article}
\usepackage[]{acl}
\usepackage{acl}
\usepackage{hyperref}
\usepackage[hang,flushmargin]{footmisc}
\usepackage{times}
\usepackage{latexsym}
\usepackage[T1]{fontenc}
\usepackage[utf8]{inputenc}
\usepackage{microtype}
\usepackage{lipsum}
\usepackage{etoolbox}
\usepackage{graphicx}
\usepackage{float}
\usepackage{lipsum}
\usepackage{amsmath}
\usepackage{enumitem}
\usepackage{multirow}
\usepackage{todonotes}
\usepackage{subfig}
\usepackage{booktabs}
\usepackage{graphicx}

\DeclareGraphicsExtensions{.pdf,.ai,.jpg,.png}
\IfFileExists{/home/stein/.signature}{}{%
\setkeys{Gin}{pagebox=artbox}
}
\graphicspath{{./}}

\newcommand{\bsfigure}[3][scale=1.0]{%
  \begin{figure}[tb]
    \centering
    \includegraphics[#1]{#2}
    \caption{#3}\label{#2}
    \vskip-1ex
  \end{figure}}

\usepackage{tabularx}
\usepackage{color}
\usepackage{multirow}
\usepackage{booktabs}
\usepackage{graphbox}
\newcommand{\Ni}{(1)~}
\newcommand{\Nii}{(2)~}
\newcommand{\Niii}{(3)~}
\newcommand{\Niv}{(4)~}

\RequirePackage{type1cm}
\RequirePackage{color}
\RequirePackage{soul}
\setstcolor{blue}
\usepackage{xcolor}
\usepackage{tikz}
\usepackage{soul}
\setulcolor{cyan}

\newcommand{\bscom}[3][]{%
  \noindent
  \st{#2}{\fontsize{10}{11}\selectfont
    \color{blue}#3\ifx\\#1\\\else{\color{violet}[#1]}\fi
    }
  }

\usepackage{newfloat}
\DeclareFloatingEnvironment[fileext=lop]{photo}

\begin{document}

\title{\mbox{\kern-12.5pt Differential Bias: On the Perceptibility of Stance Imbalance in Argumentation}\hspace*{-12.5pt}}

\newcommand{\authorspace}{\hspace{1.0em}}
\author{
Alonso Palomino$^1$
\authorspace
Martin Potthast$^1$ 
\authorspace
Khalid Al-Khatib$^2$  
\authorspace
Benno Stein$^3$  \\[1ex]
$^1$ Leipzig University\quad
\texttt{<first>.<last>@uni-leipzig.de} \\
$^2$ University of Groningen\quad
\texttt{khalid.alkhatib@rug.nl} \\
$^3$ Bauhaus-Universität Weimar\quad
\texttt{benno.stein@uni-weimar.de}
}

\date{}

\maketitle

\begin{abstract}
Most research on natural language processing treats bias as an absolute concept: Based on a (probably complex) algorithmic analysis, a sentence, an article, or a text is classified as biased or not. Given the fact that for humans the question of whether a text is biased can be difficult to answer or is answered contradictory, we ask whether an ``absolute bias classification'' is a promising goal at all. We see the problem not in the complexity of interpreting language phenomena but in the diversity of sociocultural backgrounds of the readers, which cannot be handled uniformly: To decide whether a text has crossed the proverbial line between non-biased and biased is subjective. By asking ``Is text~$X$ more [less, equally] biased than text~$Y$?'' we propose to analyze a simpler problem, which, by its construction, is rather independent of standpoints, views, or sociocultural aspects. In such a model, bias becomes a preference relation that induces a partial ordering from least biased to most biased texts without requiring a decision on where to draw the line. A prerequisite for this kind of bias model is the ability of humans to perceive relative bias differences in the first place. In our research, we selected a specific type of bias in argumentation, the stance bias, and designed a crowdsourcing study showing that differences in stance bias are perceptible when (light) support is provided through training or visual aid.
\end{abstract}

\section{Introduction}

Bias is a multifaceted phenomenon that can be wittingly or unwittingly introduced into language. \citet{walton:1999} traces the term's history with argumentation and concludes that bias implies one-sidedness. \citet{vanlaar:2007} distinguishes two forms of biased argumentation, namely the exclusion of arguments of the pro or con position (``stance bias'') and the exclusion of arguments discussing a certain aspect (frame) that is relevant to an issue. Figure~\ref{partial-order-one-sidedness} illustrates the spectrum of stance balance from a one-sided (left) to a balanced (right) situation. In between, an argument may be perceived as one-sided (biased) despite the inclusion of arguments from the other side if no sufficient balance is maintained. Nevertheless, a cleverly chosen stance imbalance can also serve as a rhetorical device for persuasion \cite{walton:1999}. Regardless of the actual balance of their arguments, both sides may accuse the other of bias due to one-sidedness.

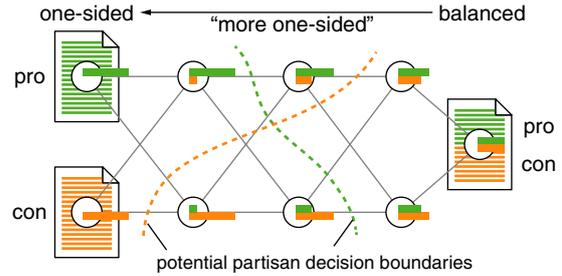
\bsfigure{partial-order-one-sidedness}{Stance bias induces a partial ordering of argumentations, the extremes being purely one-sided (left) or perfectly balanced (right). Where supporters of one side or the other draw the line between biased and non-biased depends on their degree of conviction.}

The diversity of sociocultural backgrounds, environmental conditioning, or educational attainment makes it difficult to treat bias as an absolute concept for binary classification, except for extreme cases. Granted, when focusing on the relation between texts regarding their bias, one takes a step back both in terms of problem difficulty and predication. But, trying to model the ``differential nature of bias'' (differential bias) is a valid strategy to eliminate individual, subjective factors. By developing a measure of argumentative balance with respect to the stance (stance balance) that induces a partial ordering of argumentative texts, a gradual preference relation as shown in Figure~\ref{partial-order-one-sidedness} is established. As a consequence, the evaluation of one-sidedness cannot be decided for a single text but is constrained to insights from answers to the question: ``Is text~$X$ more [less, equally] one-sided than text~$Y$?''

Using stance as an example, we investigate for the first time the extent to which differential bias is perceptible to human annotators---an important prerequisite for practical applications of debate technologies. Based on a model differential stance bias (Section~\ref{part3}), 720~human preference judgments are collected in a carefully designed crowdsourcing study (Section~\ref{part4}). Our analysis of the judgments shows that extreme imbalance is perceptible, whereas more subtle imbalance is not unless (mild) training is provided (Section~\ref{part5}).%
\footnote{\mbox{Code and data: \url{https://github.com/webis-de/AACL-22}}}
This result is important for annotating language bias in general, for argument search engines, and for developing curricula for argument analysis (Section~\ref{sum}).

\section{Related Work}
\label{part2}

After a brief review of research on bias in natural language processing~(NLP), we survey bias and one-sidedness in argumentation, and the application of pairwise judgments in corpus construction.

\subsection{Bias in Natural Language Processing}

\citet{blodgett:2020} survey 146~papers that study various forms of bias, finding that ``quantitative techniques for measuring or mitigating `bias' are [often] poorly matched to their motivations.'' \citet{sheng:2021} survey 90~papers on societal biases in language generation that tackle gender, profession, race, religion, and sexuality, among which gender bias stands out as the most frequently studied form of bias. The literature review of 61~papers by \citet{sun:2019} focuses explicitly on recognizing and mitigating gender bias in~NLP, concluding that the subfield still lacks a shared understanding, standardization, as well as evaluations that demonstrate the generalizability of current techniques. \citet{shah:2020} study bias formally and focus on how and where it is introduced into an NLP~pipeline (e.g., semantic bias in embeddings, label and selection bias in data sources). Their survey of 93~papers overviews suggested countermeasures. \citet{bender:2018} proposes the use of ``data statements'' as a means to raise awareness of ethical issues among authors, which the ACL~board has meanwhile taken up, and \citet{hovy:2016} outlines the ethical implications and impacts NLP~systems have on society.

\enlargethispage{0.5\baselineskip}
Although (social) bias has attracted much attention, to the best of our knowledge, only \citet{spliethoever:2020} explicitly study social biases in argumentation, showing that current argument corpora are biased ``in favor of male people with European-American names.'' However, two types of non-social biases have been studied more in-depth because of their relevance for argumentation: cognitive biases and one-sidedness bias.

\subsection{Cognitive Biases in Argumentation}

\citet{huang:2012} show that exposing decision makers---who exert confirmation bias via selective reading habits---to counterarguments will improve their decision outcomes. \citet{wright:2017} propose a visual ``argument mapping'' aid for intelligence analysts, which organizes arguments and counterarguments to better manage cognitive biases such as confirmation bias, anchoring, framing effects, and neglect of probability. These biases have recently also been studied by \citet{kiesel:2021b}, who discuss the challenges and opportunities of developing argumentative conversational search engines. For educators who teach argument evaluation, \citet{diana:2019,diana:2020} develop a measure that predicts if student assessments of argument strength are affected by confirmation bias; the measure is based on the alignment of individuals' values with values in political arguments. \cite{amorim:2018} find that confirmation bias impacts peer-evaluation of student essays, which then propagate into automatic essay scoring systems.

\subsection{One-Sidedness Bias in Argumentation}

\enlargethispage{-\baselineskip}
The following studies are of particular relevance to our contributions because of their focus on perceptions of bias in argumentation. \citet{walton:1999} and \citet{vanlaar:2007} argue that bias in argumentation implies one-sidedness because a biased argumentation typically fails to be balanced, favoring one side of a topic, aspect, or stance over others. One-sidedness is a phenomenon that influences the perception of information seekers: \citet{schlosser:2011} and \citet{chen:2016} investigate the relation between reviewers' expertise, their bias, and product type when evaluating the utility of online product feedback. The authors observe that customers typically consider one-sided reviews more helpful than two-sided reviews because, in one-sided reviews, users usually consider aspects such as the expertise of the reviewer to be more significant in their purchase decision. \citet{wolfe:2013} analyze the process of students in understanding one-sided arguments: In two experiments that focus on measuring reading times and analyzing how students summarize neutral texts, the authors find that the processing of one-sided arguments is based on a ``belief bias.'' \citet{kienpointner:1997} study one-sidedness of argumentation to understand the antagonistic climate of political debates about political assylum. They examine which aspects are included or excluded (global bias) and what strategies are used to address certain aspects (local bias).

Operationalizations of one-sidedness in argumentation have been contributed by \citet{potthast:2018g} and \citet{kiesel:2019c}, who study writing-style-based approaches in order to detect hyperpartisan news; they organized a shared task that received more than 40~submissions. \citet{stab:2016} choose a semantics-based approach to predict the presence or absence of opposing arguments in an argumentative text. Both approaches target only the extreme cases of one-sidedness (i.e., only pro or only con arguments; see Figure~\ref{partial-order-one-sidedness} on the far left). More generally, \citet{kucuk:2021} formulates the problem (without its operationalization) of predicting the ratio of pro, con, and neutral stances in argumentative corpora. Our work complements theirs by first answering the question whether stance bias can be perceived at~all.

\subsection{Relative Judgments in Argumentation}

\citet{kienpointner:1997} point out that argument bias assessment requires judgment in relation to the argumentative context because of their complex properties. Lacking an objective reference, another piece of argumentation can be used instead. Two recent studies follow this approach: \citet{habernal:2016} propose the task of predicting the convincingness of arguments, where given a pair of arguments with the same stance on an issue, the more convincing one must be chosen. From many such comparisons a global ranking can be statistically derived. \citet{gienapp:2020a} apply the same approach to annotate argument quality while minimizing the number of comparisons required.

Beyond argumentation, \cite{howcroft:2020} employ a relative judgment of summaries to rank them based on quality criteria, such as fluency and readability. \citet{simpson:2019} utilize pairwise comparisons to infer labels for humorous and metaphoric texts, and \citet{gooding:2019} do so to annotate words. \citet{cattelan:2012} survey such analyses beyond NLP and computer science.

\section{Measuring Differential Bias Perception}
\label{part3}

To measure the perceptibility of differential stance bias in argumentation in a controlled environment, we develop a basic experiment as shown in Figure~\ref{differential-bias-experiment-example}. For two argumentations,~$X$ and~$Y$, on the same topic, the central question to participants is: ``Which of the two argumentations is more biased?'' With a corpus of topic-labeled arguments at hand, the construction of pairs of argumentations for a given topic is straightforward. First, the number of pro arguments for~$X$ and~$Y$ is selected, each ranging from zero to four arguments. A corresponding number of pro arguments are randomly selected and randomly assigned to the available slots. Then, the remaining slots are randomly filled with random con arguments on the same topic. The two argumentations are shown side by side to be read one after the other and then compared. No 1:1~correspondence is to be assumed between arguments at the $i$-th~position, $i = 1,\ldots,4$, and therefore arguments are not aligned across argumentations.

\bsfigure{differential-bias-experiment-example}{Which of the two argumentations ($X$~vs.~$Y$, left vs.\ right) is more biased? Answer is given below.}

\begin{figure*}[t]
\includegraphics{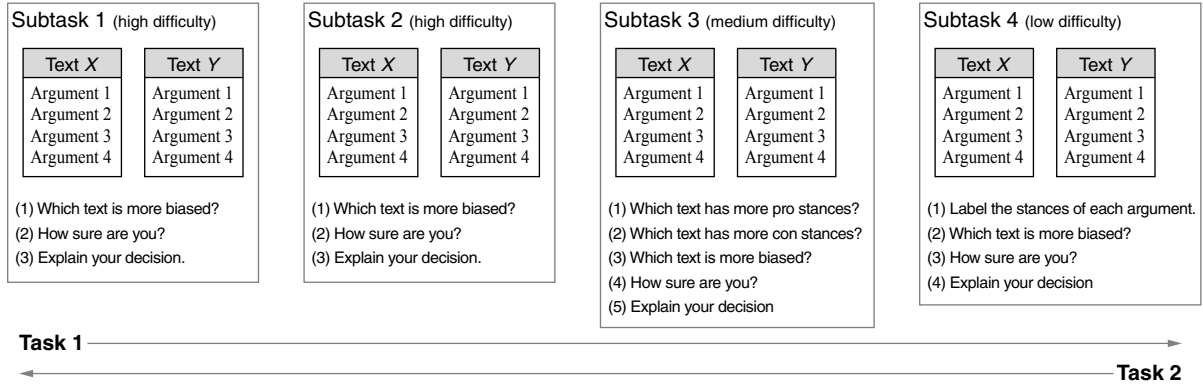}
\caption{Two tasks organize the experiments in decreasing and increasing order of difficulty, respectively.}
\label{differential-bias-hit-setups}
\end{figure*}

\enlargethispage{-2\baselineskip}
Alternative experimental setups were explored as part of a pilot study to find the simplest (``atomic'') setup to investigate our experiment variables of interest. In the following, an overview these variables is given ($D$\,=\,dependent, $I$\,=\,independent, $C$\,=\,controlled, $U$\,=\,uncontrolled)\,:
\setlength{\leftmargini}{1.5em}
\begin{itemize}
\setlength{\itemsep}{-1ex}
\item[$D$~]
perceptibility of differential stance bias 
\vspace{1ex}
\item[$I_1$~]
differential stance bias: 3~levels
\item[$I_2$~]
difficulty: low, medium, high
\item[$I_3$~]
participant training: autodidact vs.\ trained
\item[$I_4$~]
participant expertise: stance labeling accuracy
\item[$I_5$~]
participant confidence: 3-level self-assessment
\vspace{1ex}
\item[$C_1$]
argumentation length: 4~arguments each
\item[$C_2$]
argument length: 2-3~sentences each
\item[$C_3$]
argument frames: all arguments from 1~frame
\item[$C_4$]
order effects~I: random order of arguments
\item[$C_5$]
order effects~II: no textual coherence
\item[$C_6$]
opinion diversity: 9~participants per topic
\vspace{1ex}
\item[$U_1$]
stance perceptibility: e.g., explicit vs.\ implicit
\item[$U_2$]
other language biases: e.g., at the lexical level
\end{itemize}
\setlength{\leftmargini}{2.5em}

Key to measuring~$D$, the perceptibility of differential stance bias, is a model for independent variable~$I_1$: Let~$X$ denote an argumentation on a given topic that argues pro or con a proposition, where~$X = X^+ \cup X^-$ is the union of a set of pro arguments~$X^+$ and con arguments~$X^-$, so that~$X^+ \cap X^-=\emptyset$. The stance balance~$\delta$ between pro and con arguments in~$X$ is measured as the absolute size difference between~$X^+$ and~$X^-$:
\[
\delta(X) = \left||X^+| - |X^-|\right|.
\]

The closer~$\delta$ is to~$0$, the more balanced the two sides are in~$X$. Given a second argumentation~$Y$, $|X| = |Y|$, the differential stance bias can then be quantified as the absolute difference between the stance balance scores of~$X$ and~$Y$:
\[
\Delta(X,Y) = |\delta(X) - \delta(Y)|\,.
\]

The closer~$\Delta$ is to~$0$, the less strong the differential bias is between~$X$ and~$Y$. Given the designated argumentation length of $|X| = |Y| = 4$, the image of both~$\delta$ and~$\Delta$ is~$\{0,2,4\}$.

Consider the ``Tidal Energy'' example in Figure~\ref{differential-bias-experiment-example}: Argumentation~$X$ is an extreme case with only con arguments, argumentation~$Y$ is the balanced case with two pro (first, fourth) and two con arguments (second, third). This yields~$\delta(X) = 4$ and~$\delta(Y) = 0$ and thus~$\Delta(X,Y) = 4$, the maximum possible differential bias in this setting. It follows that~$X$ is more biased than~$Y$. Suppose~$Y$ consists of only pro arguments instead~($\delta(Y) = 4$)\,: Although~$X$ and~$Y$ argue exclusively on either the pro or the con side, both argumentations are equally imbalanced, so that~$\Delta(X,Y) = 0$. The intermediate case of~$\Delta(X,Y) = 2$ results when~$Y$ includes three pro and a single con argument or vice versa.

Regarding independent variable~$I_2$ (see Figure~\ref{differential-bias-hit-setups}), high difficulty means no indication is given about argument stance or stance balance before asking which of the two argumentations is more biased. Medium difficulty means we first ask which argumentation contains more pro arguments and which contains more con arguments before asking to select the more biased one---a subtle hint at stance balance. Low difficulty means the stance of each argument has to be labeled first as pro, neutral, con, or unknown, coloring each argument according to the chosen label. In this way, stance is strongly emphasized when answering the question which argument is more biased.

Experiments are presented in either decreasing (Task~1) or increasing (Task~2) order of difficulty ($I_3$; see Figure~\ref{differential-bias-hit-setups}). Task~2 is also preceded by an explanation of the importance of stance balance in relation to bias, as well as an admission test for argument stance labeling that must be successfully completed. Note the different learning experiences participants have as they move through the two tasks. Two groups of participants are distinguished~($I_4$): those who succeed in stance labeling and those who get it wrong more than once. Also, participants are asked to self-assess their confidence~($I_5$).

\begin{table*}
\small

\renewcommand{\tabcolsep}{0pt}
\renewcommand{\arraystretch}{1.2}
\begin{tabular}[t]{@{}ll@{}}
(a) \\
\toprule
\bfseries Labels & \bfseries Argument \\
\midrule
\parbox[t]{3.1cm}{\raggedright
\parbox[t]{1cm}{Topic:}  Abortion \\
\parbox[t]{1cm}{Stance:} Pro \\
\parbox[t]{1cm}{Frame:}  Fetus rights}
& \parbox[t]{0.4\textwidth}{\raggedright
No individual has rights over another individual. Therefore, a fetus cannot be said to have an inviolable right to a woman's body and sustenance from that body. A woman can, therefore, decide to deprive the fetus of the usage of her body (abortion). A fetus cannot have a right to a woman's body to sustain its life.} \\
\addlinespace
\midrule
\addlinespace
\parbox[t]{3.1cm}{\raggedright
\parbox[t]{1cm}{Topic:}  Death penalty \\
\parbox[t]{1cm}{Stance:} Con \\
\parbox[t]{1cm}{Frame:}  Internat. law}
& \parbox[t]{0.4\textwidth}{\raggedright
The U.N. does not support the death penalty. In all the courts we have set up (U.N. officials) have not included death penalty. United Nations oppose the death penalty.} \\
\addlinespace
\midrule
\addlinespace
\parbox[t]{3.1cm}{\raggedright
\parbox[t]{1cm}{Topic:}  Capitalism vs. \\
\parbox[t]{1cm}{~} Socialism \\
\parbox[t]{1cm}{Stance:} Pro \\
\parbox[t]{1cm}{Frame:}  Rights}
& \parbox[t]{0.4\textwidth}{\raggedright
This type of helpful framework neglects appeals for human rights and any other framework of deontology, morality, ethics, etc. Capitalism can embrace the utilitarian framework while not precluding any form of decision calculus in policymaking to protect human rights. Socialist leadership cannot protect human rights effectively.} \\
\addlinespace
\bottomrule
\end{tabular}
\hfill
\renewcommand{\tabcolsep}{8pt}
\renewcommand{\arraystretch}{0.95}
\begin{tabular}[t]{@{}cc@{\hspace{1em}}cc@{}}
\multicolumn{1}{@{}l}{(b)} \\[0.3ex]
\toprule
\multicolumn{2}{@{}c@{\hspace{1em}}}{\bfseries Differential Bias} & \multicolumn{2}{c@{}}{\bfseries Experiments} \\
\cmidrule(r@{1em}){1-2}\cmidrule(l){3-4}
\addlinespace[2pt]
$\Delta(X,Y)$ & $\delta(X){:}\delta(Y)$ & HITs & Workers \\
\midrule
0 & 0:0       & 6 & 54 \\
0 & 2:2       & 7 & 63 \\
0 & 4:4       & 6 & 54 \\
\midrule
2 & 2:0 / 0:2 & 7 & 63 \\
2 & 4:2 / 2:4 & 7 & 63 \\
4 & 4:0 / 0:4 & 7 & 63 \\
\bottomrule
\end{tabular}
\par
\vspace{-30ex}\hfill
\renewcommand{\tabcolsep}{9pt}
\renewcommand{\arraystretch}{0.95}
\begin{tabular}[t]{@{}cccc@{}}
\multicolumn{1}{@{}l}{(c)} \\[0.6ex]
\toprule
\bfseries Bias & \multicolumn{2}{@{}c@{}}{\bfseries Precision} & \bfseries Accuracy \\
\cmidrule(r@{\tabcolsep}){1-1}\cmidrule(l@{\tabcolsep}r@{\tabcolsep}){2-3}
\addlinespace[2pt]
$\delta(X){:}\delta(Y)$ & Pro & Con \\
\midrule
0:0     & 0.80 & 0.90 & 0.84 \\
2:2     & 0.80 & 0.66 & 0.77 \\
4:4     & 0.61 & 0.40 & 0.62 \\
\midrule
4:2 / 2:4 & 0.44 & 0.35 & 0.44 \\
4:0 / 0:4 & 0.80 & 0.62 & 0.76 \\
2:0 / 0:2 & 0.52 & 0.60 & 0.55 \\
\midrule
Avg.    & 0.66 & 0.59 & 0.66 \\[-0.5ex]
\bottomrule
\end{tabular}

\caption{
(a)~Examples of arguments from the top three most frequent topics in the Webis-Argument-Framing-19 corpus.
(b)~The stance distributions and the number of instances considered for the crowdsourcing study.
(c)~The crowd worker precision and accuracy for labeling pro/con arguments.}
\label{table-corpus-examples-and-labeling}

\end{table*}

The variables $C_1$-$C_6$ are tightly controlled by setting a specific value or by introducing diversity through randomization. Text lengths~($C_1$ and~$C_2$) are kept as short as possible, and each argument has a similar length to the others, so that confounding by length differences is avoided.$C_3$~specifically avoids mixing up stance bias with \citeauthor{vanlaar:2007}'s second type of one-sidedness in argumentation with respect to topic frames. $C_5$~is due to the requirement to generate argumentative texts rather than reusing existing ones, although the texts are generally not extremely incoherent either. Variables beyond our control include the difficulty of distinguishing pro and con stances in the argument corpus (some arguments are more subtly pro or con than others;~$U_1$), and (social) language biases that may be due to word choice~($U_2$). Nevertheless, we manually checked for hate speech and removed corresponding arguments, as well as analyzed for biased terminology using dictionaries.

\section{Crowdsourcing Study}
\label{part4}

Our crowdsourcing study implements the assessment of the perceptibility of differential stance bias on Amazon Mechanical Turk~(MTurk). We give an overview of the task design and its instantiation using an argument corpus.

\subsection{Argument Corpus}

Arguments from the Webis-Argument-Framing-19 corpus \cite{ajjour:2019b} were used to populate the tasks. It includes 12,000~arguments on 465~topics sourced from \url{debatepedia.org}. Each argument is labeled with its topic, pro or con stance, and frame. Table~\ref{table-corpus-examples-and-labeling}a shows three examples. Topics with fewer than 8~arguments of comparable length available with the same frame label were excluded~($C_2,~C_3$). From the remaining topics, 40~argument pairs were randomly generated (see Figure~\ref{differential-bias-experiment-example}). Table~\ref{table-corpus-examples-and-labeling}b shows the distribution of argument pairs across the six differential bias cases implied by our model~($I_1$).

\enlargethispage{-\baselineskip}
To exercise ``limited control'' over possible linguistic biases~($U_2$), we use the Linguistic Inquiry and Word Count~(LIWC)~2022 \cite{tausczik:2010},%
\footnote{\href{https://www.liwc.app/}{https://www.liwc.app/}}
a dictionary-based tool for identifying psycholinguistic, thematic, and tonal properties of language. All selected arguments score low on tone and swearing. Nevertheless, we manually reviewed each argument for quality and to ensure that they do not contain offensive language.

\subsection{Study Execution}

We created 20~``Human Intelligence Tasks''~(HITs) on MTurk, 10~for each of the two tasks, and distributed the 40~argument pairs to them as subtasks (see~Figure~\ref{differential-bias-hit-setups}). A total of 571~workers participated, each of whom was allowed to participate only once. Repeated participation would have undermined the learning experience of a task and thus the results achieved. One task took 45~minutes and was paid at 6~USD or 8~USD per hour.

The tasks are implemented in the form of a dialog, one subtask per step. For quality control (see below) JavaScript code was inserted to monitor the workers. As in Figure~\ref{differential-bias-experiment-example}, a pair of arguments was presented side by side in columns as titled boxes with borders. The dialog could only be worked through forward, using JavaScript to ensure that workers complete a subtask first before moving on. Moving back was prevented retrospective revisions, which is particularly relevant for Task~1, where subtasks are arranged in decreasing order of difficulty~($I_2$). Since nine workers~($C_6$) worked independently on the same subtask, the arguments in each argumentation~$X$ and~$Y$ were randomly shuffled for each worker~($C_4$).

The following measures were taken for quality control:
\Ni
work times: exclusion of workers who took less than 3~minutes to read instructions or less than 8~minutes to complete a subtask;
\Nii 
MACE score~\cite{hovy:2013}: exclusion of workers whose score is below~80\%;
\Niii
approval rate: exclusion of workers whose MTurk approval rate was below~90\%; and
\Niv
language proficiency: exclusion of workers who submitted many grammatical or spelling errors or random text in the mandatory comment fields for decision rationales in each subtask.
For Task~2, an admission test had to be passe. Finally, we ensured that each task contained subtasks with~($\Delta(X,Y)=0$) and without~($\Delta(X,Y)>0$) differential stance bias. In this way, workers who are able to distinguish the two types can be distinguished from those who cannot.

\section{Results and Analysis}
\label{part5}

\begin{table*}[ht]

\small
\centering

\renewcommand{\tabcolsep}{1pt}
\renewcommand{\arraystretch}{1.5}
\begin{tabular}[t]{@{}cc@{\hspace*{1em}}cccc@{\hspace*{1em}}ccccc@{}}
\multicolumn{1}{@{}l}{(a)} \\
\toprule
\multicolumn{2}{@{}c@{\hspace*{1em}}}{\bfseries Differential Bias} & \multicolumn{9}{c}{\bfseries Accuracy per Subtask and Difficulty} \\
\cmidrule(r@{1em}){1-2}\cmidrule{3-11}
$\Delta(X,Y)$~~ & $\delta(X){:}\delta(Y)$ & 1.1 & 1.2 & 1.3 & 1.4 &  & 2.1 & 2.2 & 2.3 & 2.4 \\
\midrule
0 &    0:0    & 0.27 & 0.33 & 0.58 & 0.66 & & 0.60 & 0.55 & 0.42 & 0.50 \\
0 &    2:2    & 0.33 & 0.28 & 0.56 & 0.57 & & 0.62 & 0.67 & 0.56 & 0.54 \\
0 &    4:4    & 0.11 & 0.36 & 0.44 & 0.64 & & 0.70 & 0.64 & 0.52 & 0.50 \\
\midrule
2 & 2:0 / 0:2 & 0.41 & 0.60 & 0.55 & 0.68 & & 0.88 & 0.86 & 0.62 & 0.70 \\
2 & 4:2 / 2:4 & 0.44 & 0.44 & 0.55 & 0.65 & & 0.77 & 0.69 & 0.66 & 0.59 \\
4 & 4:0 / 0:4 & 0.45 & 0.55 & 0.68 & 0.70 & & 0.88 & 0.66 & 0.75 & 0.66 \\
\bottomrule
\end{tabular}
\hfill
\renewcommand{\tabcolsep}{2pt}%
\renewcommand{\arraystretch}{1.4}
\begin{tabular}[t]{@{}lrcc@{\hspace*{2em}}lrcc@{}}
\multicolumn{1}{@{}l@{}}{(b)} \\[-0.8ex]
\multicolumn{8}{@{}r@{}}{\includegraphics{plot-accuracy-distribution-per-setup}} \\[0.5ex]
\multicolumn{1}{@{}l@{}}{(c)} \\
\toprule
\multicolumn{8}{@{}c@{}}{\bfseries Average Accuracy per Task and Subtask Difficulty} \\
\midrule
Sec. &\kern-1em Type & Acc. & $\Delta$\,Acc. & Sec. &\kern-1em Type & Acc. & $\Delta$\,Acc. \\
\cmidrule(r@{2em}){1-4}\cmidrule{5-8}
1.1\,/\,1.2 & 1 & 0.38 & 0.00 & 2.1         & 3 & 0.74 & \phantom{$-$}0.00 \\
1.3         & 2 & 0.55 & 0.17 & 2.2         & 2 & 0.67 &          $-$0.07 \\
1.4         & 3 & 0.64 & 0.09 & 2.3\,/\,2.4 & 1 & 0.58 &          $-$0.09 \\
\bottomrule
\end{tabular}

\renewcommand{\tabcolsep}{3pt}
\renewcommand{\arraystretch}{1.3}
\begin{tabular}{ccllllcccccclllllcccc}
\multicolumn{1}{@{}l@{}}{(d)} \\
\toprule
\multicolumn{2}{c}{\textbf{Differential Bias}} & \multicolumn{19}{c}{\textbf{Accuracy per Worker Competence Level}} \\
\cmidrule{3-21} 
\multicolumn{2}{c}{} & \multicolumn{9}{c}{Task~1} & \textbf{} & \multicolumn{9}{c}{Task~2} \\
\cmidrule(r@{\tabcolsep}){3-11}\cmidrule(l@{\tabcolsep}){13-21}
\multicolumn{2}{c}{} &
\multicolumn{4}{c}{Low} &
\multicolumn{1}{c}{} &
\multicolumn{4}{c}{High} &
\multicolumn{1}{c}{} &
\multicolumn{4}{c}{Low} &
\multicolumn{1}{c}{} &
\multicolumn{4}{c}{High} \\ 
\midrule
\multicolumn{1}{c}{$\Delta(X,Y)$} &
\multicolumn{1}{c}{$\delta(X){:}\delta(Y)$} &
\multicolumn{1}{c}{1.1} &
\multicolumn{1}{c}{1.2} &
\multicolumn{1}{c}{1.3} &
\multicolumn{1}{c}{1.4} &
\multicolumn{1}{c}{} &
\multicolumn{1}{c}{1.1} &
\multicolumn{1}{c}{1.2} &
\multicolumn{1}{c}{1.3} &
\multicolumn{1}{c}{1.4} &
\multicolumn{1}{c}{} &
\multicolumn{1}{c}{2.1} &
\multicolumn{1}{c}{2.2} &
\multicolumn{1}{c}{2.3} &
\multicolumn{1}{c}{2.4} &
\multicolumn{1}{c}{} &
\multicolumn{1}{c}{2.1} &
\multicolumn{1}{c}{2.2} &
\multicolumn{1}{c}{2.3} &
\multicolumn{1}{c}{2.4} \\
\midrule
0      & 0:0         & 0.11 & 0.00 & 0.18 & 0.28 & \multicolumn{1}{l}{} & 0.37 & 0.55 & 0.66 & 0.53 &           & 0.55 & 0.44 & 0.33 & 0.24 &  & 0.61 & 0.63 & 0.55 & 0.66 \\
0      & 2:2         & 0.14 & 0.33 & 0.42 & 0.11 &                      & 0.44 & 0.22 & 0.52 & 0.55 &           & 0.43 & 0.25 & 0.22 & 0.12 &  & 0.70 & 0.65 & 0.50 & 0.55 \\
0      & 4:4         & 0.00 & 0.26 & 0.33 & 0.24 &                      & 0.68 & 0.44 & 0.66 & 0.48 &           & 0.50 & 0.33 & 0.48 & 0.44 &  & 0.62 & 0.55 & 0.66 & 0.58 \\
2      & 2:0 / 0:2   & 0.20 & 0.44 & 0.55 & 0.35 &                      & 0.72 & 0.66 & 0.63 & 0.68 &           & 0.66 & 0.60 & 0.50 & 0.54 &  & 0.80 & 0.75 & 0.58 & 0.72 \\
2      & 4:2 / 2:4   & 0.48 & 0.45 & 0.42 & 0.28 &                      & 0.55 & 0.77 & 0.60 & 0.66 &           & 0.52 & 0.57 & 0.55 & 0.33 &  & 1.0  & 0.88 & 0.70 & 0.66 \\
4      & 4:0 / 0:4   & 0.55 & 0.36 & 0.48 & 0.60 &                      & 0.66 & 0.62 & 0.74 & 0.76 &           & 0.61 & 0.46 & 0.56 & 0.20 &  & 0.75 & 0.68 & 0.68 & 0.64 \\
\midrule
\multicolumn{2}{c}{{\color[HTML]{000000} Average}} &
\multicolumn{1}{l}{0.25} &
\multicolumn{1}{l}{0.31} &
\multicolumn{1}{l}{0.40} &
\multicolumn{1}{l}{0.31} &
\multicolumn{1}{c}{} &
\multicolumn{1}{c}{0.57} &
\multicolumn{1}{c}{0.54} &
\multicolumn{1}{c}{0.63} &
\multicolumn{1}{c}{0.61} &
\multicolumn{1}{c}{} &
\multicolumn{1}{l}{0.54} &
\multicolumn{1}{l}{0.44} &
\multicolumn{1}{l}{0.40} &
\multicolumn{1}{l}{0.29} &
\multicolumn{1}{c}{} &
\multicolumn{1}{c}{0.74} &
\multicolumn{1}{c}{0.69} &
\multicolumn{1}{c}{0.61} &
\multicolumn{1}{c}{0.63} \\
\bottomrule
\end{tabular}

\caption{
Perception accuracy of differential stance bias 
(a)~dependent on subtask, and subtask difficulty (high:~1.1,~1.2,~2.1, and~2.2; medium:~1.3 and~2.3; low:~1.4 and~2.4),
(b)~dependent on task,
(c)~dependent on task and subtask difficulty, and
(d)~dependent on worker competence at argument stance labeling.}
\label{table-crowdsourcing-results}

\end{table*}

This section reports workers' perception accuracy for differential stance bias with respect to the independent variables discussed in Section~\ref{part3}.

\subsection{Perceptibility of Differential Stance Bias}

Table~\ref{table-crowdsourcing-results}a shows workers' overall perception accuracy of differential stance bias~($I_1$). Argument pairs without differential stance bias~$\Delta(X,Y) = 0$) are less well perceived as such, with an average of~0.42 in Task~1 and~0.57 in Task~2, than argument pairs with differential stance bias~($\Delta(X,Y)>0$; average of~0.55 in Task~1 and 0.72~in Task~2). The highest accuracy is obtained for the 2:0\,/\,0:2~distribution (0.56~and 0.76~in Tasks~1 and~2) and the lowest for the 0:0~distribution (0.46~and 0.51~in Tasks~1 and~2). The higher a differential stance bias, the easier it is for workers to perceive it.

Furthermore, we find that workers who are successful at labeling the stance of individual arguments~($I_4$; part of subtasks of low difficulty) are also more successful at perceiving differential stance bias. Table~\ref{table-corpus-examples-and-labeling}c shows the workers' precision and accuracy for stance labeling. Only the pro and con arguments are displayed. Workers who could not decide (label ``unknown'') or label an argument as neutral are excluded from further analysis, since all arguments are either pro or con. An average accuracy of~0.62 is achieved for stance labeling, ranging from~0.44 to~0.84 depending on the differential bias distribution. Table~\ref{table-crowdsourcing-results}d shows that workers with high competence in stance labeling (accuracy~$>0.7$) achieve a perception accuracy for differential stance bias of~0.59 in Task~1 and~0.67 in Task~2, but workers with low competence achieve only~0.31 in Task~1 and~0.42 in Task~2.

An examination of whether workers' educational attainment (collected by questionnaire) affects perception accuracy is negative. The highest degrees attained are high school~(20\%), bachelor~(60\%), and master or higher~(20\%), and the Pearson correlations range from~-0.08 to~0.0 for Task~1 and~0.01 to~0.0 for Task~2.

\subsection{Dependence on Difficulty and Training}

Table~\ref{table-crowdsourcing-results}c shows perception accuracy as a function of difficulty~($I_2$) and difficulty order~($I_3$). In both tasks, perception accuracy increases the lower the difficulty of the subtask. The accuracy at high difficulty is~0.38 for Task~1 and~0.58 for Task~2. Accuracy at medium difficulty is~0.55 for Task~1 (0.17~improvement over high difficulty) and~0.67 for Task~2 (0.09~decrease below high difficulty). At low difficulty, workers achieve the highest accuracy of~0.64 in Task~1 and~0.74 in Task~2. Although an improvement in perception accuracy can be observed between high and low difficulty, workers in Task~1 seem to be confused by the change from high to medium difficulty. However, the relative improvement in accuracy between high and low difficulty is the same for both tasks.

The perception accuracy differs in absolute terms between Task~1 (high to low difficulty) and Task~2 (training, then low to high difficulty). Table~\ref{table-crowdsourcing-results}b shows that workers achieve an accuracy of~0.38 to~0.60 with an average of~0.52 in Task~1 and an accuracy of~0.51 to~0.76 with an average of~0.64 in Task~2. This accuracy difference (about~0.15) suggests that there is a positive effect of training workers to perform more difficult subtasks, rather than expecting them to detect differential stance bias off hand. The difference is significant according to Welch's $t$-test, with a test statistic of~4.28 and a $p$ value of~0.001.

\subsection{Worker Confidence and Feedback}

Participants had to self-assess their confidence in each subtask as ``very sure,'' ``reasonably sure,'' and ``uncertain''~($I_5$). When workers gave either of the first two self-assessments, they have higher perception accuracies, albeit dependent on of the presence of differential stance bias. For Task~1, an accuracy of~0.58 is achieved in cases with differential stance bias, and an accuracy of~0.69 for Task~2. The lowest accuracies are observed when workers self-assess their confidence as ``not sure'', where the accuracy falls to~0.42 in Task~1 and~0.65 in Task~2.

\begin{table*}[t]
\small
\centering
\renewcommand{\tabcolsep}{5pt}%
\renewcommand{\arraystretch}{1.2}%
\begin{tabular}{@{}cc>{\raggedright\arraybackslash}p{2.1cm}>{\raggedright\arraybackslash}p{4.68cm}>{\raggedright\arraybackslash}p{4.68cm}@{}}
\toprule
\multicolumn{2}{c}{\bfseries Differential Bias} & \multicolumn{3}{c}{\bfseries Decision Justification} \\
\cmidrule(r{\tabcolsep}){1-2}\cmidrule(l{\tabcolsep}){3-5}
$\Delta(X,Y)$ & $\delta(x):\delta(y)$ & Topic
& Correct
& Incorrect
\\

\midrule

0 & 0:0     & Globalization
& I'm not completely sure about the second paragraph in text X but I think overall both texts have 2 con and 2 pro paragraphs: X:~pro-con-con-pro Y:~con-pro-con-pro
& It is a complex issue and at least one argument on both texts could be constructed as in favor or against globalization, depending on the reader's initial stance.
\\

\midrule

0 & 2:2     & Adult Incest
& Both texts have pros and cons regarding incest.
& Text Y features three pro arguments, whereas text X has an equal share of pro and con arguments.
\\

\midrule

0 & 4:4     & Animal Testing
& Both X and Y have an even amount of pro and con arguments. The first argument in Y is in a slight gray area, but it seems to be against the spirit of animal testing overall so. I counted it as a con.
& X favors the side of animal life being sacred  more and a more emotional argument chain
\\

\midrule

2 & 2:0 / 0:2 & Banning  Cell Phones  in Cars
& Y posts three points on the good cellphones in cars can do and though it does present the cons they are vastly outweighed in the argument
& Because X is more pro arguments
\\

\midrule

2 & 4:2 / 2:4 & Tidal Energy
& Text Y is more one-sided as it explains more cons of tidal energy than pros of it, whereas text X has an equal number of pros and cons.
& Y is talking about  megawatts and energy and anchors and cables while X speaks in a Reader's Digest form that most are able to understand
\\

\midrule

4 & 4:0 / 0:4 & Atheism
& X only makes arguments against atheism and tries to justify acts of religion. Y gives quite logical arguments and doesn't seem to be very one sided.
& X seems more to add their own opinion, while Y explains more of the facts.
\\

\bottomrule
\end{tabular}
\caption{Examples of worker decision justifications dependent on differential bias.}
\label{table-labeling-justifications}
\end{table*}

For a qualitative assessment of the workers' abilities, we asked them to justify their decision as to which of the two argumentations is more biased. In reviewing the collected justifications, we found that some workers cited other types of bias in addition to stance bias. Table~\ref{table-labeling-justifications} shows selected justifications of the workers on subtasks with different differential bias distributions, depending on whether the workers chose the more biased argumentation correctly, or incorrectly. While one worker argues about the emotional tone of the argument chain on a subtask with differential bias distribution of~4:4, another worker addresses the factuality of the arguments on a subtask with a distribution of~4:0/0:4. Some workers base their decisions on the perceived degree of subjectivity or objectivity of their arguments. Others argue about the constructiveness, persuasiveness, rationality, and verifiability of the arguments. Comparing the justifications in Task~1 with those in Task~2, many workers in Task~1 stick to their initial explanations used in the first subtasks. In contrast, workers in Task~2 more often realize that the study is about stance balance.

\section{Conclusions and Future Work}
\label{sum}

How one-sided can an argumentation be without this one-sidedness going unnoticed? In political debates, parties often accuse each other of one-sidedness. An advanced debating skill hence is to subtly shift the balance of an argumentation so that it goes unnoticed, especially by potential voters. But political debate is perhaps only one of the most extreme cases in which skewed argumentation can be found, as \citet{kienpointner:1997} point out in their in-depth analysis of letters to the editor regarding the topic of political asylum. We believe that a perfectly balanced argument is the exception rather than the rule because, after all, the goal is to convince the audience of one's own opinion. For this reason, audiences must try to be alert to (subtle) manipulation opportunities, and here formal training to detect differences in stance bias seems a promising future direction. The results of our study indicate that differences in stance bias are perceptible, and even more so when (as in our crowdsourcing setting) workers are trained with a visual aid in the form of green/red signals that emphasize the stance of the arguments. In an online setting where arguments are read at leisure, a tool that highlights the distribution of stances could help readers make more informed decisions, not to mention formal training in the context of a debate.

Our study further demonstrates, to our knowledge for the first time, the viability of assessing bias via relative judgments rather than absolute ones. In this regard, we focus on stance bias for its ease of operationalization. It is likely that many other types of bias can be similarly addressed in future work, although not all of them can be operationalized as easily as in our study. In any case, the obvious benefits of reducing the subjective bias of annotators certainly justify further investigation and the development of a theory in this direction. This should, of course, include research on the interactions of biases, since not all biases can be easily controlled. For instance, some workers based their judgments not only on differences in stance, but also pointed to other language properties that, in their opinion, made one argumentation seem more one-sided than another. Although we took great care to exclude examples of language bias, subtle biases may still have been present in our sample. A straightforward and fascinating application of our differential bias model is to extend it to the study of frame bias.

\section*{Ethical Statement}

Our study involved human annotators recruited via the Amazon Mechanical Turk~(MTurk) crowdsourcing platform. In accordance with fair compensation guidelines for crowd workers, we ensured an appropriate hourly rate. As is common with Mturk, some worker accounts attempted to submit fake data. We reviewed all cases that met our study's exclusion criteria regarding rejection of submitted work. As part of the quality control process, in addition to open rejection, we also introduced ``internal rejection,'' which resulted in acceptance of the submitted work with full payment but exclusion from our analysis. We decided to play it safe and discard internally in ambiguous cases. All workers were informed about the study and agreed to the quality controls and compensation conditions before work began.

One limitation of our study is that it focuses on differences between stance biases. We are aware that several factors play a role in making an argumentation biased such as the frame of an argumentation, logical connections between arguments, (lack of) cohesion and strength of arguments, or social biases resulting from loaded language. However, it remains an open problem to define the semantics of these biases. Of course, laboratory experiments are always limited in this or similar ways, and our experiment is no different. We do not foresee any unethical uses of our results or its underlying tools, but hope that it will contribute to advancing the discourse on bias, and help to depart from from absolute bias claims.

\bibliography{aacl22-differential-bias-lit}

\end{document}